\documentclass[10pt,twocolumn,letterpaper]{article}

\usepackage[pagenumbers]{cvpr} % To force page numbers, e.g. for an arXiv version

\definecolor{cvprblue}{rgb}{0.21,0.49,0.74}
\usepackage[pagebackref,breaklinks,colorlinks,allcolors=blue]{hyperref}

\usepackage[vlined,linesnumbered,ruled]{algorithm2e}

\def\paperID{*****} % *** Enter the Paper ID here
\def\confName{CVPR}
\def\confYear{2025}

\title{4th PVUW MeViS 3rd Place Report: Sa2VA}

\author{
 Haobo Yuan$^{1}$, Tao Zhang$^{2}$, Xiangtai Li$^{2}$, Lu Qi$^{2}$, Zilong Huang$^{2}$, Shilin Xu$^{2}$,
 \\
 Jiashi Feng$^{2}$, Ming-Hsuan Yang$^{1}$ \\
  {\normalsize {$^{1}$UC Merced}  \quad  {$^{2}$Bytedance} } \\
}

\begin{document}
\maketitle
\begin{abstract}
Referring video object segmentation (RVOS) is a challenging task that requires the model to segment the object in a video given the language description. 
MeViS is a recently proposed dataset that contains motion expressions of the target objects, leading to a challenging benchmark, compared with existing RVOS benchmarks.
% which requires the model to have a holistic understanding of the whole video. 
%
On the other hand, for referring expression tasks, a new trend is to adopt multi-modal large language model (MLLM) to achieve better image and text alignment.
In this report, we show that with a simple modification to the test time inference method on stronger MLLMs, we can lead to stronger results on MeVIS.
In particular, we adopt the recent method Sa2VA, a unified model for dense grounded understanding of both images and videos.
By enlarging the scope of key frames, \textbf{without} any further training, we can achieve the 3rd place in the 4th PVUW workshop.
% we propose Sa2VA, a unified model for dense grounded understanding of both images and videos to tackle this scenario. Our method achieves 
%
Our code is available at \url{https://github.com/magic-research/Sa2VA}.
\end{abstract}    
\section{Introduction}
\label{sec:intro}

Referring video object segmentation (RVOS) is a challenging task that aims to segment and track objects in the video according to the language expression. 
MeVIS~\cite{MeViS} is a referring video object segmentation dataset focused on motion expressions driven video segmentation, which is more challenging than datasets focused on appearance expression driven video segmentation, such as Ref-DAVIS~\cite{khoreva2019video} and Ref-YTVOS~\cite{seo2020urvos}. 
The motion expression-driven video object segmentation requires models to have fine-grained understanding abilities of videos, including both object \textit{appearance} and \textit{motion}, as well as good video object segmentation capabilities.

Recently, multimodal large models (MLLMs)~\cite{Qwen-VL, Qwen2-VL, Qwen2.5-VL, chen2024internvl, chen2024far, chen2024expanding, zhou2025they} have demonstrated very powerful image and video understanding capabilities, including understanding of overall sense, comprehension of object appearance and actions, and understanding of relationships between objects. 
The video segmentation foundation model SAM-2~\cite{ravi2024sam} has achieved performance and generalization capabilities far exceeding previous video segmentation methods~\cite{zhang2023dvis, zhang2025dvis++, zhou2024dvis, li2023tube, xu2024rap, zhang20231st_pvuw, zhang20231st_lsvos} through its powerful data engine. 
Some grounded MLLMs~\cite{lai2024lisa, zhang2024omg} have proven that good instruction-driven segmentation can be achieved by combining MLLMs and segmentation experts~\cite{kirillov2023segment, li2024omg, yuan2024mamba}. 
Based on these priors, Sa2VA~\cite{yuan2025sa2va} combines the SOTA MLLM InternVL2.5~\cite{chen2024expanding} and SAM-2~\cite{ravi2024sam} to create a powerful grounded MLLM, demonstrating strong image and video understanding and segmentation capabilities.

In this challenge, we adopt Sa2VA~\cite{yuan2025sa2va} and optimize its frame sampling strategy. 
Specifically, Sa2VA's original inference setting is to directly use the first 5 frames of the video as input, which is unreasonable because the first 5 frames contain very limited object motion information, creating a huge challenge for motion expression aware video object segmentation. 
To solve this problem, we expand the frame sampling interval from 1 to 3, which can encompass a longer time range to help Sa2VA more accurately identify object motion, thereby improving performance on MeVIS~\cite{MeViS}.

Without any finetuning on specific datasets, test augmentation, or model ensembling, Sa2VA-26B achieves 56.3 J$\&$F on the competition. 
Finally, we obtain third place in the competition, demonstrating the powerful potential of grounded MLLMs.

\section{Method}
\label{sec:method}

\begin{figure*}[t]
  \centering
  \includegraphics[width=.95\linewidth]{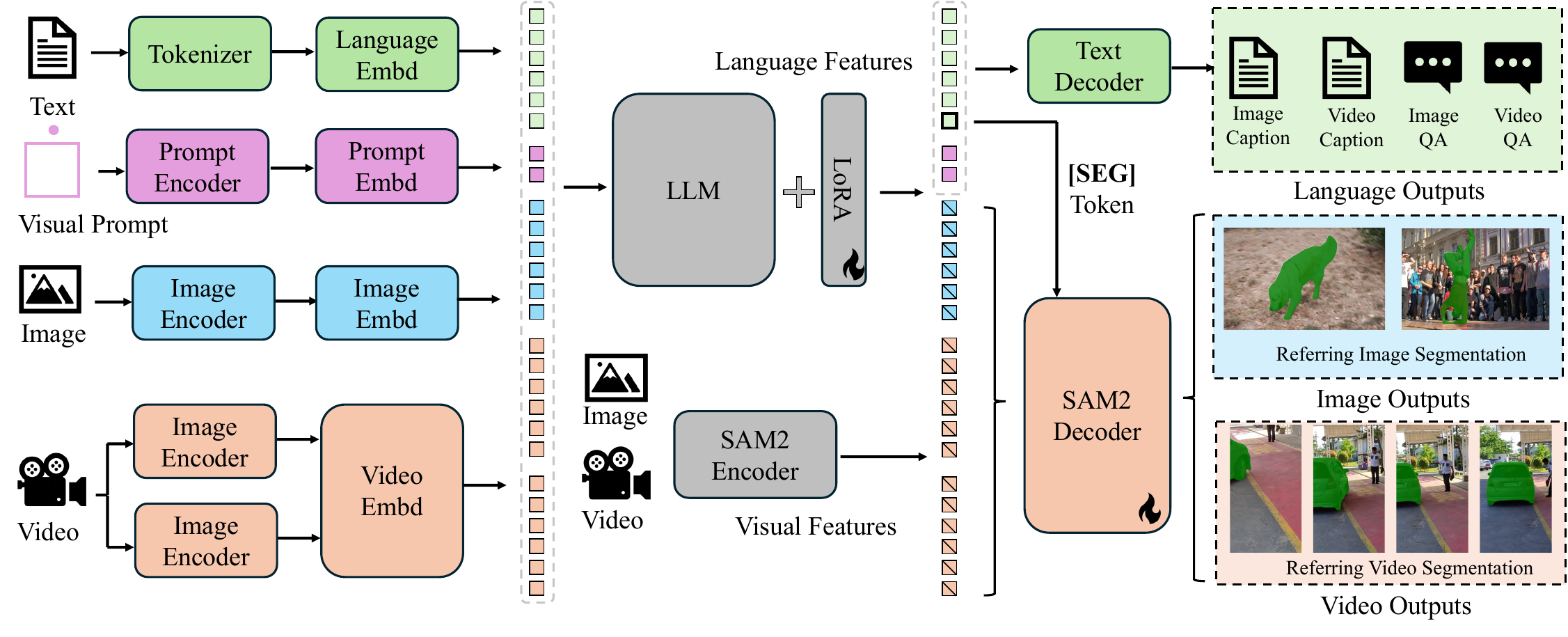}
  \caption{\textbf{The Sa2VA model.} The model first encodes the input texts, visual prompts, images, and videos into token embeddings. These tokens are then processed through a large language model (LLM). The output text tokens are used to generate the ``[SEG]'' token and associated language outputs. The SAM-2 decoder receives the image and video features from the SAM-2 encoder, along with the ``[SEG]'' token, to generate corresponding image and video masks. Modules with a {red}{fire} icon are trained during the one-shot instruction-tuning.
  Note that we do not train the model for MeVIS dataset and we only adopt pre-trained model~\cite{yuan2025sa2va} for inference. }
  \label{fig:method}
\end{figure*}

In this section, we will first introduce our baseline model, Sa2VA~\cite{yuan2025sa2va} in Sec.~\ref{sec:method_sa2va}, and we introduce the detailed modification on the inference pipeline for the MeViS dataset in Sec.~\ref{sec:method_test_time}.

\subsection{Sa2VA}
\label{sec:method_sa2va}

\noindent\textbf{Meta Architecture.} As shown in Fig.~\ref{fig:method}, Sa2VA consists of an MLLM and SAM2. The MLLM accepts inputs of images, videos, and text instructions, and outputs text responses based on the text instructions. When the user instruction requires the model to output segmentation results, the text response will include the segmentation token ``[SEG]". The segmentation token's hidden states serve as implicit prompts and are converted through SAM2 into image and video-level object segmentation masks.

\noindent\textbf{MLLM.} The SOTA MLLM InternVL 2.5~\cite{chen2024expanding} is adopted as the MLLM, demonstrating powerful capabilities in single-image, multi-image, and video understanding and conversation. 
InternVL 2.5 adopts a LLaVA-like~\cite{liu2023visual} architecture, consisting of an InternVIT~\cite{chen2024internvl}, an MLP projector, and a Large Language Model. High-resolution images are first divided into several sub-images and a thumbnail, then encoded by InternVIT into vision tokens, which are mapped through one MLP and combined with text tokens as input to the LLM. The LLM will autoregressively output text responses, which may include segmentation tokens. The segmentation token's hidden states from the last LLM transformer layer are processed through an MLP to serve as the prompt input for SAM2~\cite{ravi2024sam}.

\noindent\textbf{SAM2.} SAM2 generates object segmentation results for some high-resolution video frames based on the segmentation prompts output by the MLLM. Subsequently, SAM2 propagates these frame segmentation results to obtain object segmentation results for the entire video.

\noindent
\textbf{Sa2VA Model Training.} The original Sa2VA is co-trained on multiple datasets, including image/video VQA datasets, caption datasets, and image/video referring segmentation datasets, including MeViS. For this challenge, we do not fine-tune the model for MeViS, where we only focus on test time modifications on Sa2VA.

\noindent
\textbf{Naive Ref-VOS Inference Pipeline.} As described in \cref{alg:refvos_inf}, the origin pipeline of Sa2VA begins by extracting the first five frames ($k_1$, $k_2$, $\ldots$, $k_K$ are set to 1, 2, 3, 4, and 5 respectively) of the input video as keyframes, ensuring that they capture the critical context for the following processing. 
These key frames are fed into CLIP and flattened to visual sequential tokens for LLM processing. 
The LLM takes the visual and language tokens as input and uses these tokens to extract information about the video to generate the ``[SEG]'' token. 
In SAM-2, the prompt encoder encodes boxes or clicks to prompt embeddings for object referring. 
Different from SAM-2, we use two linear layers to project the ``[SEG]'' token into the language prompt embedding, which serves as an extension of the SAM-2 prompt encoders. 
With the language prompt embedding, we use the SAM-2 decoder to generate the masks of the key frames. 
Then, we use the memory encoder of SAM-2 to generate a memory based on the output key-frame masks. 

Finally, the memory attention in SAM-2 generates the remaining masks using the memory generated from the key-frame and previous non-key-frame masks.
%

% \begin{algorithm}[!t]
% \scriptsize
% \caption{Ref-VOS Inference Pipeline}\label{alg:refvos_inf_ori}
% \textbf{Input:} Video length $N$; Number of key frames $M$; Video frames $S_{N}$ ($X_1$, $X_2$, $X_3$,$\ldots$, $X_N$); Language description $T$;\\
% \textbf{Output:} Sequence of masks $M_1$, $M_2$, $M_3$,$\ldots$, $M_N$;\\
% \textbf{Run:} Sa2VA Model for Ref-VOS;\\
% Extract key frames: $S_{M}$ $\gets$ $S_{N}$;\\
% Visual embeddings: $E_v$ $\gets$ Encoder($S_{M}$);\\
% Language embeddings: $E_l$ $\gets$ Encoder($T$);\\
% Answers: $A$ $\gets$ LLM($\{E_v, E_l\}$);\\
% Prompt embedding: $P_l$ $\gets$ Linear(Find($A$, '[SEG]'));\\
% \For{$i = 1,2,\ldots,M$}{
%     SAM-2 feature: $F_{i}$ $\gets$ Encoder($X_0$);\\
%     Mask: $M_i$ $\gets$ Decoder($\{P_l, F_{i}\}$);\\
%     Update Memory: $Mem$ $\gets$ Cross-Attention($\{Mem, M_i\}$);\\
% }
% \For{$i = M+1,M+2,\ldots,N$}{
%     SAM-2 feature: $F_{i}$ $\gets$ Encoder($X_0$);\\
%     Mask: $M_i$ $\gets$ Decoder($\{Mem, F_{i}\}$);\\
%     Update Memory: $Mem$ $\gets$ Cross-Attention($\{Mem, M_i\}$);\\
% }
% \textbf{emit} $M_1$, $M_2$, $M_3$,$\ldots$, $M_N$;
% \end{algorithm}

\subsection{Test time augmentation for Sa2VA on MeVIS}
\label{sec:method_test_time}
\noindent
\textbf{Long-Interleaved Inference.}
The Naive Ref-VOS inference pipeline directly uses the first several frames as the keyframes. However, this may lead to suboptimal performance when the initial frames lack sufficient context for accurate reference embedding. This is especially evident when the language prompt requires a longer temporal reasoning.
To address this issue, we propose an inference strategy named Long-Interleaved Inference (LII). We intentionally lengthen the time duration of the key frames to capture more context in the video. 
Specifically, we interleave keyframes across a longer temporal window rather than selecting them consecutively from the beginning. 
We sample keyframes at fixed intervals throughout the video, ensuring both early and late contextual signals are incorporated into the reference embedding.
To keep the whole method simple and not overly dependent on hyperparameters, we use the same interval in all videos.
The whole algorithm is described in the Algorithm~\ref{alg:refvos_inf}. The whole algorithm is similar to the naive Ref-VOS inference pipeline, and the main difference is the key frame selection strategy. $k_1$, $k_2$, $\ldots$, $k_K$ can be set to a fixed set of values before the execution of the entire pipeline.
With the Long-Interleaved Inference strategy, the keyframes are no longer clustered at the beginning but are spread across a longer video clip. This design encourages the model to capture long-term dependencies, which is particularly beneficial in scenarios where the object appearance or scene context changes over time.

\noindent
\textbf{Other Attempts.} We also try a model ensembling strategy during the competition, which shows performance degradation and is not adopted in the final result. For the model ensembling strategy, we use two separate SAM-2 decoders during inference. The first one is from the Sa2VA, which is trained with the one-shot instruction tuning process and different from the original SAM-2 decoder as shown in Figure~\ref{fig:method}.
The other one is from the original SAM-2. 
In the process of predicting the key frame masks, we have to use the SAM-2 decoder of Sa2VA because we need to use ``[SEG]'' token as prompt. We input the key frame masks into the second SAM-2 decoder to infer the rest of the masks.
We hope to try to use this approach to separate reasoning and tracking. However, we observe a performance drop and do not apply this strategy to maintain the performance.
We also present the results of this strategy in Section~\ref{sec:main_result}.

\begin{algorithm}[!t]
\scriptsize
\caption{MeViS dataset Inference Pipeline}\label{alg:refvos_inf}
\textbf{Input:} Video length $N$; Number of key frames $K$; Video frames $S_{N}$ ($X_1$, $X_2$, $X_3$,$\ldots$, $X_N$); Language description $T$; \textbf{Key Frame Selection Strategy}: {$k_1$, $k_2$, $\ldots$, $k_K$}. \\
\textbf{Output:} Sequence of masks $M_1$, $M_2$, $M_3$,$\ldots$, $M_N$;\\
\textbf{Run:} Sa2VA Model for Ref-VOS;\\
Extract key frames: $S_{M}$ $\gets$ \{$X_{k_1}$, $X_{k_2}$, $X_{k_3}$,$\ldots$, $X_{k_K}$\};\\
Visual embeddings: $E_v$ $\gets$ Image-Encoder($S_{M}$);\\
Language embeddings: $E_l$ $\gets$ Tokenizer($T$);\\
Answers: $A$ $\gets$ LLM($\{E_v, E_l\}$);\\
Prompt embedding: $P_l$ $\gets$ Linear(Find($A$, '[SEG]'));\\
\For{$i = 1,2,\ldots,K$}{
    SAM-2 feature: $F_{k_i}$ $\gets$ SAM-Encoder($X_{k_i}$);\\
    Mask: $M_{k_i}$ $\gets$ SAM-Decoder($\{P_l, F_{k_i}\}$);\\
    Update Memory: $Mem$ $\gets$ Cross-Attention($\{Mem, M_{k_i}\}$);\\
}
\For{$i = 1,2,\ldots,N$}{
    SAM-2 feature: $F_{i}$ $\gets$ SAM-Encoder($X_i$);\\
    Mask: $M_i$ $\gets$ SAM-Decoder($\{Mem, F_{i}\}$);\\
    Update Memory: $Mem$ $\gets$ Cross-Attention($\{Mem, M_i\}$);\\
}
\textbf{emit} $M_1$, $M_2$, $M_3$,$\ldots$, $M_N$;
\end{algorithm}
\section{Experiments}

\subsection{Implementation Details}
We directly use the Sa2VA-26B model~\cite{yuan2025sa2va} as the baseline to test the results. 
The Sa2VA-26B model starts from InternVL2.5-26B~\cite{chen2024internvl} and SAM2~\cite{ravi2024sam} models. 
The training pipeline follows the Sa2VA~\cite{yuan2025sa2va}.
Sa2VA uses a one-shot instruction-tuning process on both image and video data to train the model, which means it is a general model and therefore does not need to be trained again on this dataset. 
During the inference, we add the LII strategy to improve the performance on the longer video. Specifically, we extract the 1st, 4th, 7th, 10th, 13th frames (totally 5 frames) of each video as the key frames.

\begin{table}[t!]
    \centering
    \begin{tabular}{l|l|c|cc}
    \toprule
    Rank & Team &  J\&F &  J & F\\
    \midrule
    \#1 & MVP-Lab & 62.0 & 58.8 & 65.1 \\ 
    \#2 & ReferDINO-Plus & 60.4 & 56.8 & 64.1\\
    \rowcolor{lightgray} \#3 & Sa2VA & 56.3 & 52.7 & 59.8\\
    \#4 & Pengsong & 55.9 & 53.1 & 58.8\\
    \#5 & ssam2s & 55.2 & 52.0 & 58.3\\
    \bottomrule
    \end{tabular}
    \caption{The competition leader board of 4th PVUW MeViS challenge. There are a total of 32 teams in the competition.}
    \label{tab:leaderboard}
\end{table}

\begin{table}[t!]
    \centering
    \begin{tabular}{l|c|cc}
    \toprule
    Method &  J\&F &  J & F\\
    \midrule
    Sa2VA-26B & 54.1 & 50.5  & 57.7 \\
    \rowcolor{lightgray} Sa2VA-26B + LII & 56.3 & 52.7 & 59.8 \\
    Sa2VA-26B + SAM2 & 51.6 & 48.5 & 54.7 \\
    Sa2VA-26B + SAM2 + LII & 54.2 & 51.2 & 57.2 \\
    \bottomrule
    \end{tabular}
    \caption{Ablation study of inference strategy of Sa2VA. SAM2 refers to a model ensembling strategy (integrating another SAM2 model). \textbf{LII} refers to the \textbf{Long-Interleaved Inference} strategy.}
    \label{tab:ablation_study}
\end{table}

\begin{figure*}
    \centering
    \includegraphics[width=0.98\linewidth]{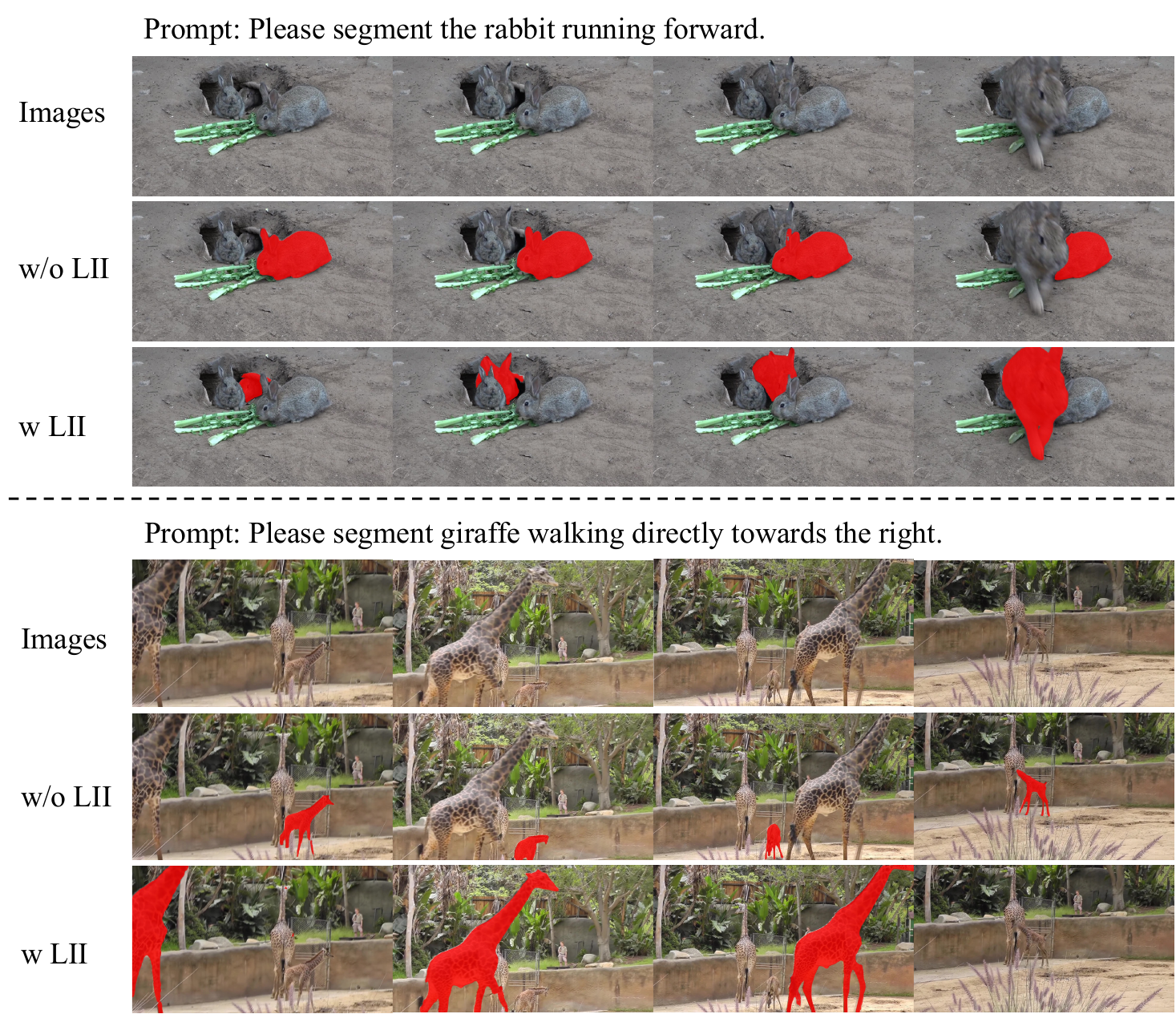}
    \caption{\textbf{Visualization comparison}. Sa2VA with Long-Interleaved Inference (LII) pipeline (i.e., w LII) shows with better understanding of the motion information in longer videos compared to without the LII pipeline (w/o LLI).}
    \label{fig:vis_comparision}
    \vspace{-3mm}
\end{figure*}

% Prompt (up): \textit{Please segment the rabbit running forward}. Prompt (bottom): \textit{Please segment the giraffe walking directly towards the right}.
\subsection{Main Results}\label{sec:main_result}
\noindent
\textbf{Competition Results.}
The final competition results are shown in Table~\ref{tab:leaderboard}. Although we do not conduct additional training, our Sa2VA-based method achieves 56.3 J$\&$F on the competition and ranks third among all 32 teams.

\noindent
\textbf{Ablation Study.}
In Table~\ref{tab:ablation_study}, we compare different inference strategies. Specifically, we evaluate the impact of using the LII strategy and model ensembling strategy. As shown in the table, the application of LLI leads to a noticeable improvement of about 2.2 J$\&$F, demonstrating the effectiveness of leveraging the longer context in the video. In contrast, the model ensembling strategy using two SAM-2 decoders does not achieve better results, and there is a performance degradation under two different settings. This may be because the introduction of a fixed module that has not been end-to-end trained cannot make good use of the knowledge in the training data. Therefore, in the final result, we do not use such a model ensembling strategy.

\noindent
\textbf{Visualization Analysis.} In Figure~\ref{fig:vis_comparision}, we present qualitative comparisons between different inference strategies to better understand their effects. The visualization results clearly show that the LII strategy enables the model to capture the context from a longer temporal range for the motion reasoning. In contrast, the baseline method often fails to capture the correct object. For example, in the first case, the prompt asks for the rabbit that moves forward. However, in the early part of the video, there is no clear clue indicating which rabbit will move forward. In this situation, the method without LII fails to localize the correct object and thus cannot perform accurate segmentation. In contrast, with the LII pipeline, the correct object can be effectively identified and segmented.

\section{Conclusion}
\label{sec:conclusion}
In this report, we explore the effectiveness of leveraging long-term context in the RVOS task. We demonstrate that the Long-Interleaved Inference (LII), which is a simple modification during the inference, can have a notable improvement even without further training of the model. Our Sa2VA with LII achieves  56.3 J$\&$F and ranks third place among 32 participating teams in the 4th PVUW MeViS competition. Our findings suggest that careful design of the inference process can lead to a notable performance gain, providing an insight for future research in RVOS.

{
    \small
    \bibliographystyle{ieeenat_fullname}
    \bibliography{main}
}


\end{document}